\documentclass[final,5p,times,twocolumn]{elsarticle}

\usepackage{amssymb}

\usepackage{multirow}
\usepackage{multicol}
\usepackage{array}
\usepackage{booktabs}
\usepackage{amsmath}
\usepackage{graphicx}
\usepackage{subfigure}
\usepackage{color}
\usepackage{tabularx}
\usepackage[colorlinks,linkcolor=blue,anchorcolor=blue,citecolor=blue,bookmarksopen,bookmarksdepth=2]{hyperref}
\usepackage{cleveref}
\crefname{figure}{Fig.}{Figs.}
\Crefname{figure}{Figure}{Figures}
\crefname{section}{Section}{Sections.}
\Crefname{section}{Section}{Sections}
\crefname{equation}{Eqn.}{Eqns.}
\Crefname{equation}{Equation}{Equations}
\crefname{table}{Table}{Tables}
\Crefname{table}{Table}{Tables}
\journal{Information Fusion}

\begin{document}

\begin{frontmatter}



\title{Remote Sensing Image Fusion Based on Two-stream Fusion Network}


\author[VRTS]{Xiangyu Liu}
\ead{xyliu@buaa.edu.cn}
\author[VRTS,DM]{Qingjie Liu\corref{mycorrespondingauthor}}
\cortext[mycorrespondingauthor]{Corresponding Author.}
\ead{qingjie.liu@buaa.edu.cn}
\author[VRTS]{Yunhong Wang}
\ead{yhwang@buaa.edu.cn}
\address[VRTS]{The State Key Laboratory of Virtual Reality Technology and Systems,\\Beihang University, Beijing 100191, China}
\address[DM]{Beijing Key Laboratory of Digital Media, School of Computer Science and Engineering, Beihang University, Beijing 100191, China}

\begin{abstract}
	Remote sensing image fusion (also known as pan-sharpening) aims at generating high resolution multi-spectral (MS) image from inputs of a high spatial resolution single band panchromatic (PAN) image and a low spatial resolution multi-spectral image. Inspired by the astounding achievements of convolutional neural networks (CNNs) in a variety of computer vision tasks, in this paper, we propose a two-stream fusion network (TFNet) to address the problem of pan-sharpening. Unlike previous CNN based methods that consider pan-sharpening as a super resolution problem and perform pan-sharpening in pixel level, the proposed TFNet aims to fuse PAN and MS images in feature level and reconstruct the pan-sharpened image from the fused features. The TFNet mainly consists of three parts. The first part is comprised of two networks extracting features from PAN and MS images, respectively. The subsequent network fuses them together to form compact features that represent both spatial and spectral information of PAN and MS images, simultaneously. Finally, the desired high spatial resolution MS image is recovered from the fused features through an image reconstruction network. Experiments on Quickbird and \mbox{GaoFen-1} satellite images demonstrate that the proposed TFNet can fuse PAN and MS images, effectively, and produce pan-sharpened images competitive with even superior to state of the arts. 
\end{abstract}
\end{frontmatter}

\section{Introduction}
Most remote sensing applications require images at the highest resolution both in spatial and spectral domains which is very hard to achieve by a single sensor. To alleviate this problem, many optical Earth observation satellites, such as QuickBird, GeoEye and IKONOS, carry two kinds of optical sensors, acquiring multi-modal data with different but complementary characteristics, in which the panchromatic sensor acquires high spatial resolution images with only a single band, while the multi-spectral sensor acquires low spatial resolution images with multiple bands. These modalities are known as panchromatic (PAN) image and multi-spectral (MS) image, respectively. The technique of PAN and MS image fusion (also known as pan-sharpening) is to fuse information from PAN and MS modalities to generate images with spatial resolution of PAN images and spectral resolution of the corresponding MS images, simultaneously. 

Pan-sharpening can be helpful for many practical  applications, such as change detection, land cover classification, so it has been raising much attention within remote sensing community. Many research efforts have been devoted to developing pan-sharpening algorithms  during the last decades~\cite{thomas2008synthesis,vivone2015critical,ghassemian2016review}. Most of these methods can be classified into three categories: 1) component substitution (CS) methods; 2) am$\acute{\mathrm{e}}$lioration de la r$\acute{\mathrm{e}}$solution spatiale par injection de structures (ARSIS) concept methods (which means enhancement of the spatial resolution by structure injections); and 3) model-based methods. The CS methods assume that the geometric detail information of a MS image lies in its structural component which can be obtained by transforming it into a new space. Then the structural component is substituted or partially substituted by the PAN to inject the spatial information. Pan-sharpening is achieved after an inverse transformation. The PCA based~\cite{chavez1991comparison,shahdoosti2016combining}, the IHS based~\cite{tu2001new,xu2014high} and the Gram-Schmidt (GS) transform~\cite{laben2000process} based methods are those of the most widely known CS methods. The fundamental assumption of the ARSIS concept methods is that the missing spatial information in MS can be inferred from the high frequencies of the corresponding PAN images~\cite{ranchin2000fusion}. To pan-sharpen a MS image, ARSIS methods apply multi-resolution algorithms, such as discrete wavelet transform (DWT)~\cite{pradhan2006estimation}, “$\grave{\mathrm{a}}$ trous” wavelet transform~\cite{nunez1999multiresolution} or curvelet transform~\cite{nencini2007remote} on a PAN image to extract high-frequency information and then inject it into a MS image. The model based methods construct degradation models of how PAN and MS images are degraded from the desired high resolution MS image and restore it from the degradation models~\cite{aly2014regularized,wei2015bayesian}. Methods beyond this scope are also developed to address the pan-sharpening problem. For instance, Li~\cite{li2011new} and Zhu~\cite{zhu2013sparse} modeled  pan-sharpening from compressed sensing theory. He et al.~\cite{he2014new} introduced a variational model based on spatial and spectral sparsity priors for pan-sharpening. Liu et al.~\cite{liu2014pan} addressed pan-sharpening from a manifold learning framework. 

Recently, deep learning techniques, especially convolutional neural networks (CNNs), have been applied to various research fields and achieved astonishing performance~\cite{dong2016image,kim2016accurate,ren2015faster,zhang2016cnn}, motivating researchers in remote sensing community to apply CNNs on pan-sharpening problems. Inspired by SRCNN ~\cite{dong2016image}, Masi et al.~\cite{PNN} proposed a pan-sharpening method based on convolutional neural networks (CNNs). They utilized a three-layered CNN architecture, which was originally designed for image super-resolution~\cite{dong2016image}, to achieve pan-sharpening. Zhong et al.~\cite{zhong2016remote} presented a CNN based hybrid pan-sharpening method, in which CNN was employed to enhance the spatial resolution of the MS image, then the GS transform was utilized to fuse the enhanced MS and PAN image to obtain the pan-sharpened images. The network used to enhance the spatial resolution of MS image is also a three-layered CNN similar to SRCNN~\cite{dong2016image}. 

Motivated by the great successes achieved by CNNs both in pan-sharpening and other research fields, and the fact that CNN is a powerful tool to extract hierarchical features of an image, we propose to perform pan-sharpening task in feature level instead of pixel level that the previous CNN based methods belong to. To achieve this end, we design an encoder-decoder like two-stream fusion network that encodes (fuses) features of PAN and MS, and then decodes the fused features to recover the pan-sharpened image. The code of our method is available at \mbox{\color{blue}{https://github.com/liouxy/tfnet\_pytorch}}.

To summarize, the main contributions of this paper are in three-fold:
\begin{itemize}
	\item We propose a two-stream CNN architecture to address the problem of pan-sharpening and accomplish it in feature level fusion. 
	\item We propose to use $\ell_1$ loss instead of widely used $\ell_2$ loss function to optimize the network. It achieves much better results than $\ell_2$.
	\item Residual learning is explored to further boost the performance of the proposed two-stream fusion network. We demonstrate that the residual learning could gain the improvement in pan-sharpening problem.
\end{itemize}

This is an extension of our work~\cite{liu2018remote}. Compare with it, this paper provides a more comprehensive and systematic report of our work. Furthermore, backgrounds of CNN and related work are added in this paper. And we provide motivations for designing the proposed network and two improvements of it to further boost the performance. Last but not least, more extensive experiments are presented to valid our methods. 

The rest of this paper is organized as follows. In \cref{sec:backgroudandrelatedwork}, we introduce background of CNNs and give a brief review on CNN based pan-sharpening methods.~\Cref{sec:tspn} explains the motivation of our two-stream pan-sharpening network and elaborate the detail architecture. ~\Cref{sec:experiments} gives experiments and comparisons with other methods. Finally, this paper is concluded  in \cref{sec:conclusion}.

\begin{figure*}[htbp]
	\centering
	\subfigure[PAN]{\label{fig:example_pan}
		\includegraphics[width=43.5mm]{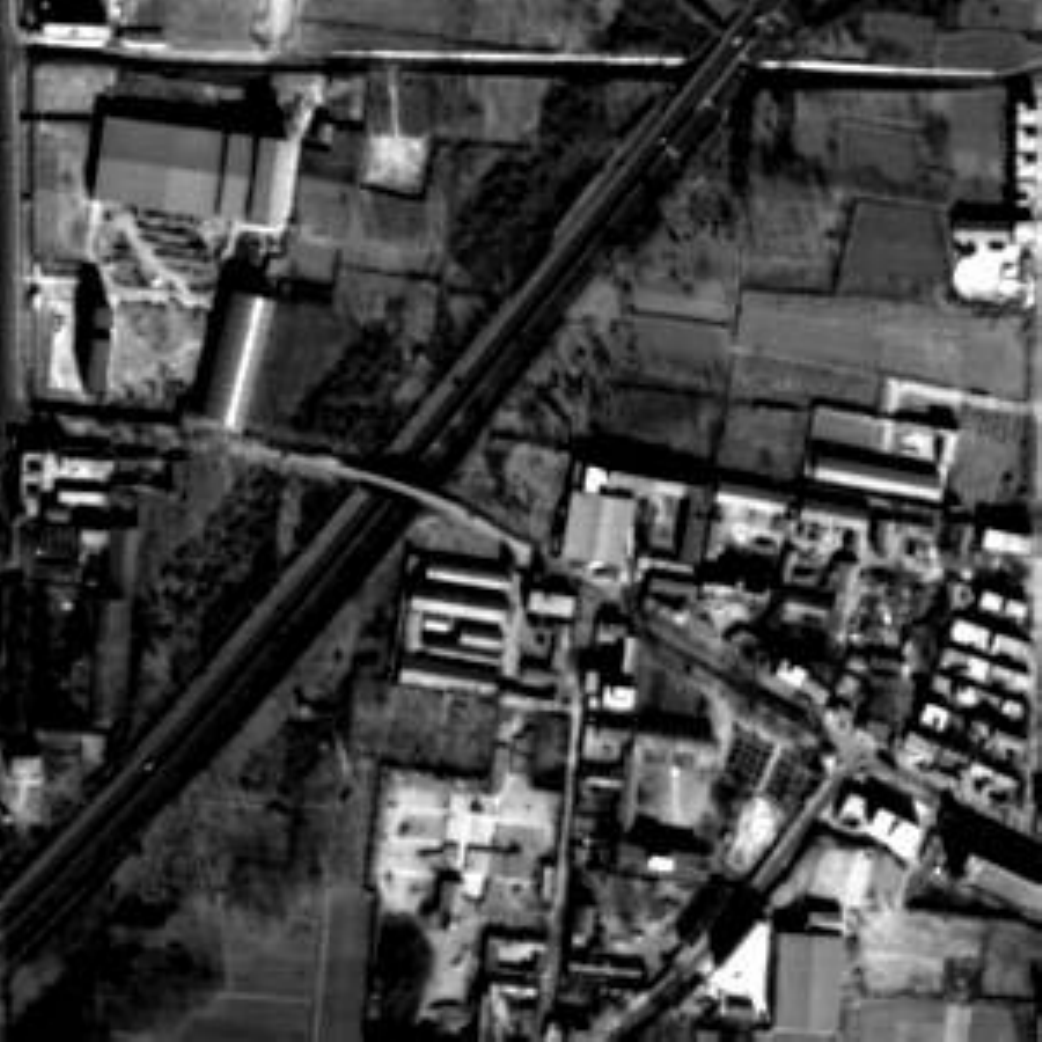}}
	\subfigure[MS]{\label{fig:example_ms}
		\includegraphics[width=43.5mm]{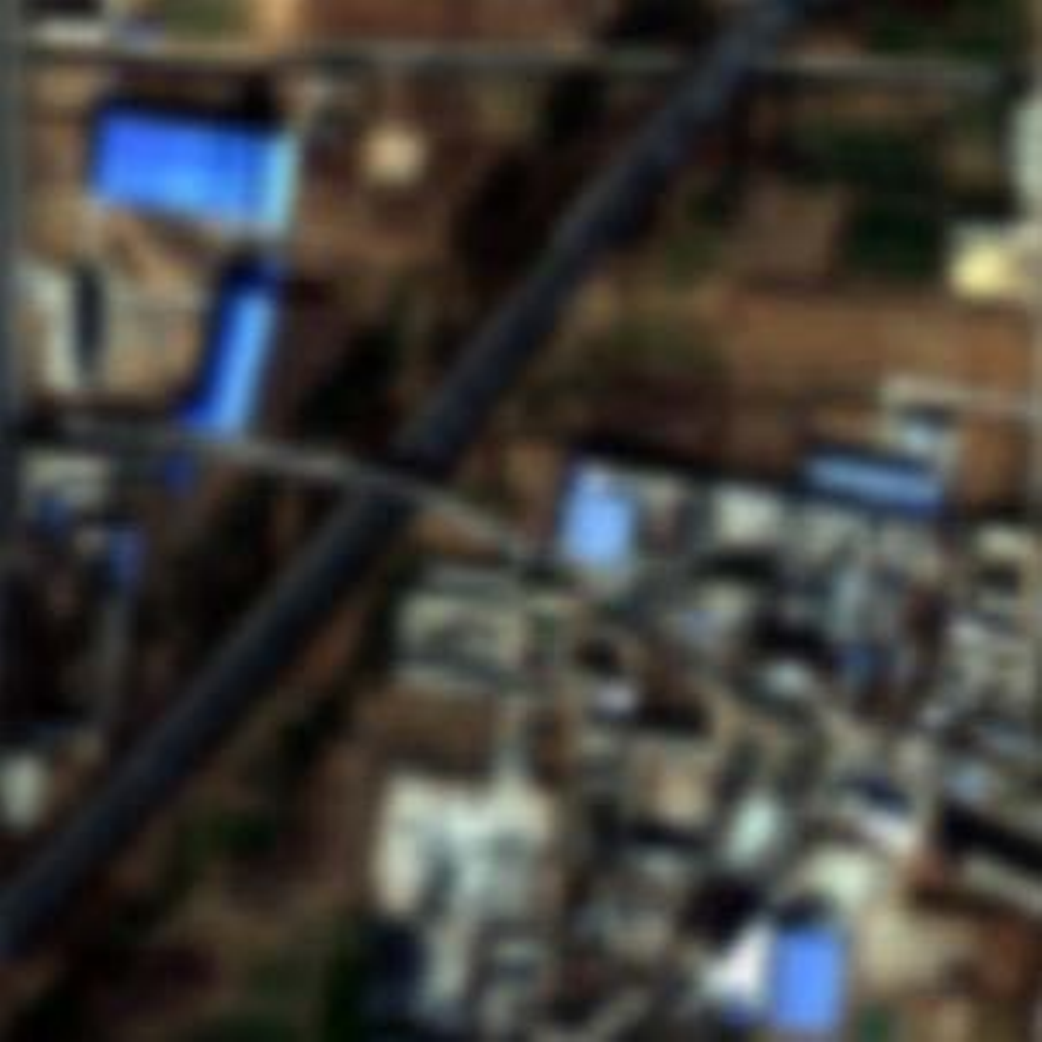}}
	\subfigure[Ground truth]{\label{fig:example_gt}
		\includegraphics[width=43.5mm]{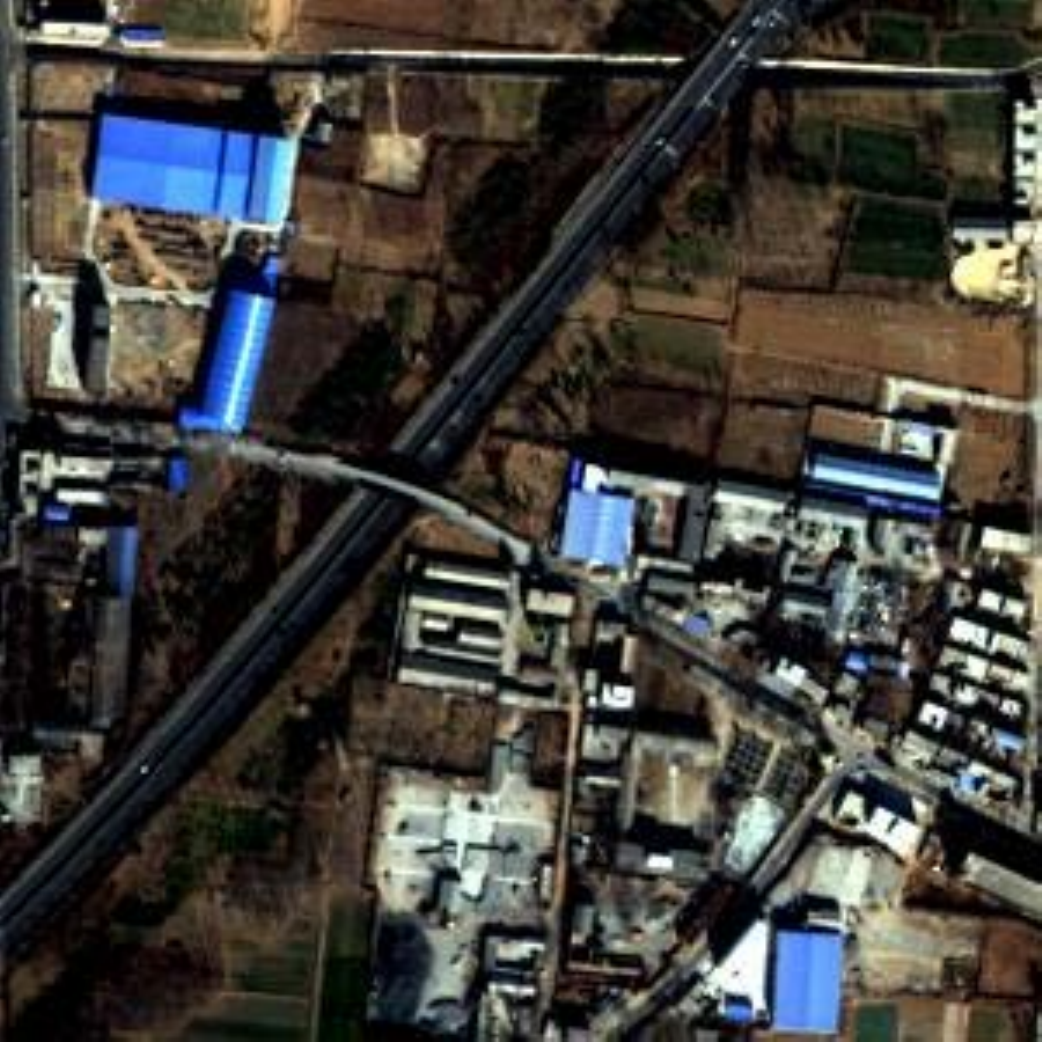}}
	\subfigure[Pan-sharpened]{\label{fig:example_tf}
		\includegraphics[width=43.5mm]{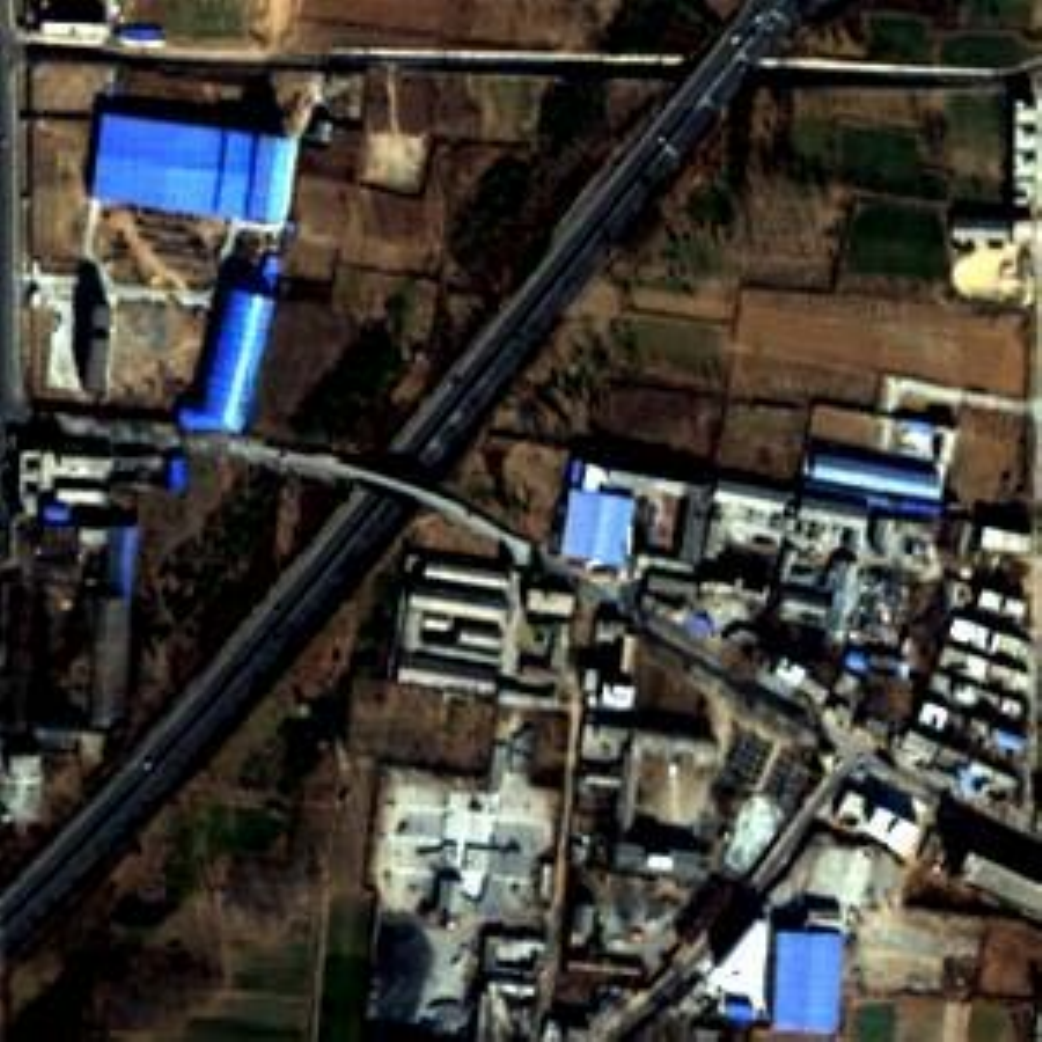}}	
	\caption{An example result of the proposed TFNet.}
	\label{fig:sample}
\end{figure*}

\section{Background and Related Work}
\label{sec:backgroudandrelatedwork}

\subsection{Convolutional Neural Networks}
Recent years have witnessed a rapidly development of deep learning techniques, which facilitate various computer vision and image processing tasks, such as image classification~\cite{krizhevsky2012imagenet}, video classification~\cite{karpathy2014large}, object detection~\cite{ren2015faster,redmon2016you}, image segmentation~\cite{long2015fully,noh2015learning}, image super-resolution~\cite{dong2016image,kim2016accurate}, multi-focus image fusion~\cite{liu2017multi,du2017image}, and pan-sharpening~\cite{PNN,zhong2016remote,rao2017residual,wei2017boosting,yang2017pannet}, in which one of the widely studied and used deep learning models is convolutional neural network (CNN). 

The design of CNN was inspired by the organization of visual cortex in mammal animals that individual cortical neuron only responds to a sub-region of visual field, called \textit{receptive field}, and neighboring neurons share similar even overlap receptive fields~\cite{hubel1968receptive}. By mimicking this discovery, a neuron in CNN is locally connected to its previous layer and replicated across the entire visual field. These characteristics, known as local connectivity and weight sharing, will dramatically reduce the number of parameters in the network, allowing the researchers to design much deeper CNN architectures with fewer parameters. In CNN, each neuron in the same visual field is realized by convolving its previous layer with a linear filter, adding a bias term and then activated by a non-linear function. The output of this neuron is called a feature map. There are multiple filters in each convolutional layer, producing a set of feature maps. Normally, a subsequent \emph{pooling} layer (\emph{max}- or \emph{mean-pooling}) is applied to down-sample the spatial size of the feature maps. Pooling operation plays an important role in CNN. By doing this, it will not only reduce the amount of parameters, and hence to avoid overfitting, but also make the learnt features invariant to shift and rotation. CNN can learn rich hierarchical representations from  training data. When an image passing trough a trained CNN, each layer, either convolutional layer or pooling layer, from low level to high level can serve as features of this image~\cite{chen2014vehicle,hou2017change}. Low level features represent local and detail information of the image, while high level features represent global and more abstract and information of the image. 

A pioneering work is LeNet-5~\cite{lecun1998gradient}, which was specially designed to recognize the handwritten characters. Although LeNet-5 achieved promising performance on handwriting recognition, CNNs are not favored by researchers due to difficulties in training deep neural networks, until the AlexNet, a breakthrough made by Krizhevsky et al.~\cite{krizhevsky2012imagenet}. AlexNet won the first place in ImageNet Large Scale Visual Recognition Challenge 2012 (ILSVRC-2012) competition. Compared with LeNet-5, several new techniques, such as dropout regularization~\cite{srivastava2014dropout} and a non-saturating activation function ReLU~\cite{nair2010rectified} were used. Dropout, which is setting the output of a hidden neuron to zero with a probability of $p$ (normally $p=0.5$) during training, is a simple yet powerful technique to prevent the network from overfitting. And the utilization of ReLU will make it several times faster to train a deep CNN. These techniques are widely adopted in follow-up CNN architectures, such as VGG nets~\cite{Simonyan14c} and GoogleNet~\cite{szegedy2015going}, to speed up the training and decrease the overfitting. 

The depth of networks is of central importance for designing CNN architectures. Deeper CNN will boost the performance, however it is almost intractable to train the network due to gradient degradation problem. Recently, He et al.~\cite{he2016deep} introduced a deep residual learning framework to harness the degradation problem, making it possible to build extremely deep residual nets. While training a CNN, optimizing each layer (or entire network) can be considered as learning a mapping function from input to output. Instead of learning the direct mapping, the residual learning divides the output into two parts, identity part, which is equal to the input, and residual part, which is the difference between output and input. Residual learning hypothesize that it is easier to optimize the residual mapping than original one~\cite{he2016deep}. By adding a shortcut connection between the input and output, a residual block is forced to learn parameters mapping the input to residuals. With this residual block, one can build network with extremely depth. It has been reported that, a lot of vision tasks can be further improved by simply replacing plain CNN with deep residual network~\cite{dai2016r,lim2017enhanced}.

\subsection{CNN for pan-sharpening}
In the field of image fusion, MS image pan-sharpening also benefits from CNN. One of the first attempts of applying CNN to pan-sharpening problem is made by Masi et al.~\cite{PNN}. In this work, pan-sharpening was regarded as a special form of image super-resolution, and achieved by a three-layered CNN with similar structure to that of super-resolution CNN (SRCNN)~\cite{dong2016image}. Since pan-sharpening takes two images, i.e. PAN and MS as inputs, while SRCNN only has one input, Masi et al. stacked the upsampled MS with PAN to form a five-band image as input. To further improve the performance, larger kernels were employed in the central layer,\footnote{They used $5 \times 5$ kernels in central layer, while in SRCNN the kernel size is $1\times 1$.} and prior knowledge was introduced to the input data to guide the learning process. Zhong et al.~\cite{zhong2016remote} also adopted SRCNN to perform pan-sharpening. The difference is that in \cite{zhong2016remote} the SRCNN was employed to enhance the resolution of MS, then GS transform was further applied on the enhanced MS and the PAN image to accomplish the pan-sharpening. 

Inspired by the success of applying residual network in low-level vision tasks, researchers in the community of pan-sharpening also explore to use the idea of residual learning to improve the performance of pan-sharpening. For instance, Rao et al.~\cite{rao2017residual} proposed a residual learning CNN model to fulfill the pan-sharpening, in which SRCNN was employed to learn the residuals between the ground truth and up-sampled MS image. The final results were obtained by adding the up-sampled MS to the predicted residuals. Similarly, Wei et al.~\cite{wei2017boosting} proposed a much deeper network to learn the residuals of the input MS and ground truth MS images. Although named residual network, these two networks are built with plain neural units. Yang et al.~\cite{yang2017pannet} introduced a deep network called PanNet for pan-sharpening. Considering that pan-sharpening is different from image super-resolution and other image fusion problems, domain-specific knowledge was incorporated when designing the PanNet. Specially, learning was performed in the high-pass filtering domain rather than the image domain to preserve spatial information, and the up-sampled MS image was added up the output to get the final results to preserve the spectral information. 
\section{Two-stream Fusion Network}
\label{sec:tspn}
\subsection{Motivation}
It is generally accepted that PAN and MS images contain \mbox{different} information. PAN image is carrier of geometric detail (spatial) information, while MS image preserves spectral information. The goal of pan-sharpening is combining the spatial details and spectral information to generate a new high resolution MS image. However, it is very hard to define what exactly spatial and spectral information are and how to represent them independently. Previous researches such as ARSIS concept methods~\cite{ranchin2000fusion,nencini2007remote} believe the spatial information lies in the high frequency of the PAN images. CS methods~\cite{APCA,AIHS} apply a linear transformation on MS image to obtain a component that contains the degraded spatial information which is then substituted by the PAN image to inject the spatial information. Recently proposed CNN based methods either considered pan-sharpening as a super-resolution problem~\cite{PNN} or using CNNs as tools to extract spatial details (i.e. residuals)~\cite{wei2017boosting,yang2017pannet}. These methods can generate results with good visual quality, however still suffer from spectral distortions or artifacts. This is mainly because it is almost impossible to extract features representing spatial or spectral information, separately. Although PAN is thought to be carrier of spatial information, there may also exists spectral information in it.

Recall that, there are three different levels of fusion strategies, namely pixel level fusion, feature level fusion and decision level fusion~\cite{ghassemian2016review}. Most of the aforementioned methods can be considered as pixel level fusion. Motivated by the fact that CNN is an exceptionally powerful tool to represent hierarchical features of an input and image can be reconstructed from these features perfectly~\cite{gatys2015texture,gatys2016image}, we propose to fuse the PAN and MS image in feature domain and then recover the pan-sharpened image from the fused features. The intuition behind this is simple, since we want to combine the advantages of the PAN and MS, the features of the pan-sharpened image should contain both information of PAN and MS. 

\subsection{Problem formulation}
Suppose $\mathcal{F}$ is a feature extractor such that the features extracted by $\mathcal{F}$ is a complete representation of an input image $I$, which means $I$ can be recovered from $\mathcal{F}(I)$ without losing any information. Most hand-crafted features, such as HOG, LBP and SIFT do not satisfy this demand. Sparse coding, which represents an image with a learned over-complete dictionary, can be used as a good feature extractor. Works have been done to develop pan-sharpening method by employing sparse representation techniques~\cite{li2011new,zhu2013sparse}.  However, fusion is still performed in image domain. It is still unclear how to fuse the features of two images in sparse domain. 

CNNs provide us a good solution to accomplish feature extraction and fusion, simultaneously. Based on above discussions, firstly we use CNNs as feature extractors to represent PAN and MS images. For convenience, in the following we use $\Phi_M(\Theta_{M})$  and $\Phi_P(\Theta_{P})$ to denote CNNs for MS and PAN, respectively. For two images $X_{P}$ and $X_{M}$ representing PAN and MS, their features extracted by CNNs can be written as $\Phi_{P}^{l}(X_{P})$ and  $\Phi_{M}^{l}(X_{M})$, where the superscript $l$ indicates that features are extracted from the $l$-th layer. After obtaining features of PAN and MS, the subsequent step is fusing them together. Because features extracted by CNNs are in the form of channel-wise feature maps, one possible way of fusion is applying \emph{pooling} operation, such as max-pooling or mean-pooling on the two feature maps. However, this will loss information, which should be avoided during pan-sharpening. Alternatively, we consider another fusion strategy by simply concatenating them together,
\begin{equation}\label{eq:concatenation}
\Phi_{f}(X_{P},X_{M}) = \Phi_{P}^{l}(X_{P}) \oplus \Phi_{M}^{l}(X_{M})
\end{equation}
where $\Phi_{f}(X_{P},X_{M})$ is the fused feature and `$\oplus$' means concatenation operation. Then the concatenated features are fed into a fusion network to form more compact representation, and followed by a reconstruction network to recover the pan-sharpened images. These stages can be integrated into one network and trained in  an end-to-end manner. Because the proposed network accepts two inputs, and thus has a two-steam architecture we call it Two-stream Fusion Network (TFNet) for simplicity.

\subsection{TFNet}
\begin{figure}[!t]
	\centering  
	\includegraphics[width=9cm]{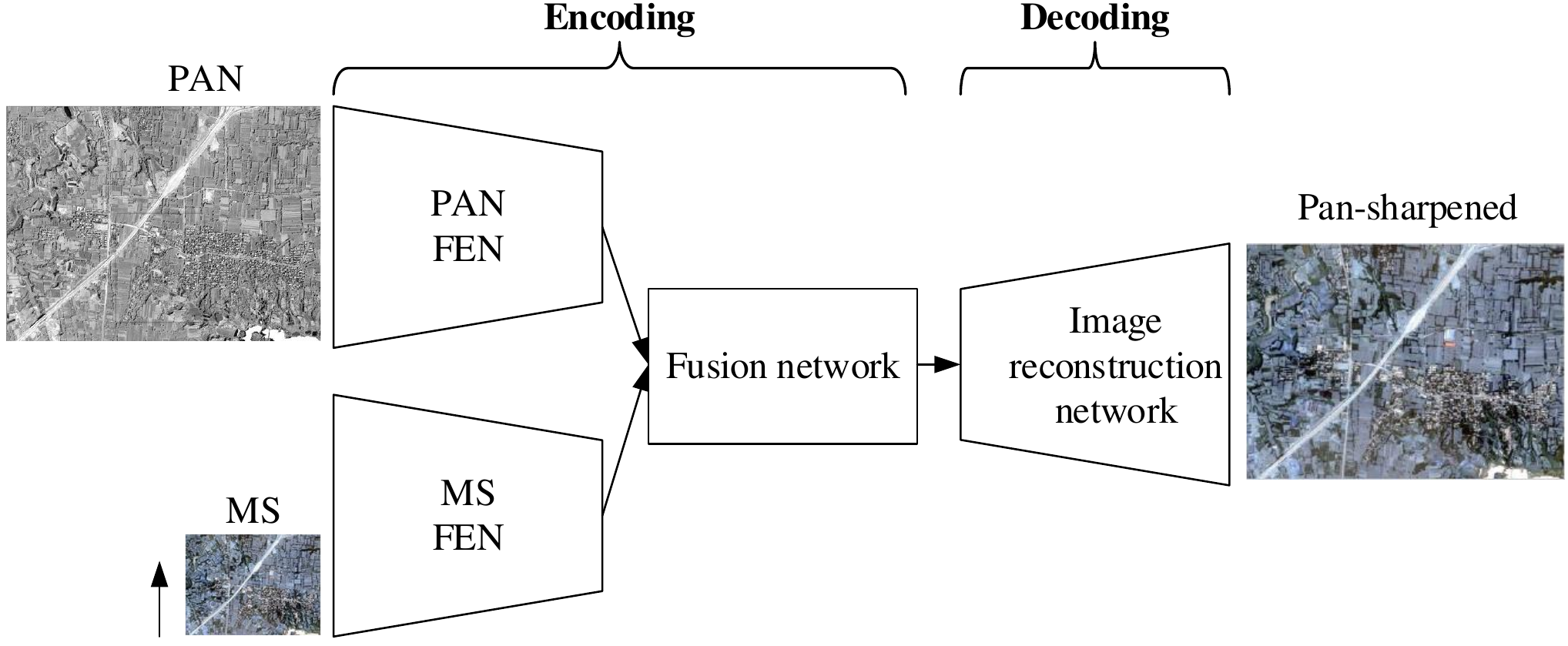} 
	\caption{Schematic illustration of the proposed TFNet. FEN is the abbreviation of feature extraction network. MS image is up-sampled to the same size of the PAN before feeding into the MS FEN.}
	\label{fig:schematicstructure}  
\end{figure}
In this subsection, we give the detail structures of our TFNet, which has an encoder-decoder like architecture and consists of three parts, including feature extraction, feature fusion and image reconstruction. The first two parts act like an encoder extracting and encoding features from the input PAN and MS images. The last part decodes the fused features to reconstruct the desired high resolution MS image. The schematic architecture of the proposed TFNet is shown in~\cref{fig:schematicstructure}.

\subsubsection{Feature extraction networks}

We use two sub-networks $\Phi_{P}$ and $\Phi_{M}$ to extract features from PAN and MS images, respectively. These two sub-networks have similar architecture but different weights. One sub-network takes a 4-band MS image as input and the other one takes a single band PAN image as input. Each of the feature extraction sub-networks consists of two successive convolutional layers followed by a Parametric Rectified Linear Unit (PReLU)~\cite{he2015delving} and a down-sampling layer. Most of the CNN architectures utilize max or mean pooling to get scale- and rotational-invariance features, however the detail information is more important in pan-sharpening, so through the entire network, we use convolutional kernels with a stride of 2 for down-sampling instead of simple pooling strategy.
\subsubsection{Fusion network}
After feature extraction, we have two volumes of feature maps $\Phi_{P}^{l}(X_{P})$ and  $\Phi_{M}^{l}(X_{M})$ representing PAN and MS images, respectively. The two feature maps explicitly capture complementary information of PAN and MS. Considering that the desired high resolution MS image should have high resolutions both in spatial and spectral domain, the features of it must capture spatial and spectral information, simultaneously. To this end, the two feature maps are concatenated together as described in~\Cref{eq:concatenation} . After that, a fusion network with three convolutional layers is applied to encode the concatenated feature maps into more compact representations. The end of the fusion network is a $[w/4, h/4, 256]$\footnote{$w$ and $h$ are the width and height of the input images, 256 is channels of the fused feature maps. The denominators in the first two elements are 4, because we down sample the feature maps two times.} tensor which encodes spatial and spectral information of the two input images. 

\begin{table*}[!htb]
	\centering
	\caption{The parameters of the TFNets. The TFNet and ResTFNet have the same parameters, except that ResTFNet has two more convolutional layers after Concat2 and Concat3 layers. This is identified in \textit{italic} in the following table.}
	\label{Table:structure}
	\renewcommand{\arraystretch}{1.2}
	\begin{tabularx}{0.9\textwidth}{p{3cm}p{3cm}XXX}
		\hline
		&    & Layer & \begin{tabular}[c]{@{}l@{}}Kernel/ Stride\end{tabular} & Output \\ 
		\hline
		\multirow{2}{*}{Input}                                                        	& MS &  &  &$128\times128\times4$\\
			\cline{2-5}
			& PAN &  &  &$128\times128\times1$\\ 
		\hline
		\multirow{6}{*}{\begin{tabular}[c]{@{}l@{}}Feature\\ Extraction\end{tabular}}    
		& \multirow{3}{*}{MS} &Conv1\_M& $3\times3\times32/1$ &$128\times128\times32$\\  
		&                     &Conv2\_M& $3\times3\times32/1$ &$128\times128\times32$\\ 
		&                     &Conv3\_M& $2\times2\times64/2$ &$64\times64\times64$\\ \cline{2-5} 
		& \multirow{3}{*}{MS} &Conv1\_P&$3\times3\times32/2$&$128\times128\times32$\\  
		&                     &Conv2\_P&$3\times3\times32/1$&$128\times128\times32$\\ 
		&                      &Conv3\_P&$2\times2\times64/2$&$64\times64\times64$\\
		\hline
		\multirow{4}{*}{\begin{tabular}[c]{@{}l@{}}Feature\\ Fusion\end{tabular}}         
		&  &Concat1&-&$64\times64\times128$\\  
		&  &Conv4&$3\times3\times128/1$&$64\times64\times128$\\ 
		&  &Conv5&$3\times3\times128/1$&$64\times64\times128$\\
		&  &DownConv6&$2\times2\times256/2$&$32\times32\times256$\\  
		\hline
		\multirow{11}{*}{\begin{tabular}[c]{@{}l@{}}Image \\ Reconstruction\end{tabular}} 
		&  &Conv7&$3\times3\times256/1$&$32\times32\times256$\\ 
		&  &Conv8&$3\times3\times256/1$&$32\times32\times256$\\
		&  &UpConv9&$2\times2\times128/2$&$64\times64\times128$\\  
		&  &Concat2&-&$64\times64\times256$\\ 
		&  &\textit{ResConv1}&$1\times1\times128/1$& \textit{$64\times64\times128$}\\
		&  &Conv10&$3\times3\times128/1$&$64\times64\times128$\\ 
		&  &Conv11&$3\times3\times128/1$&$64\times64\times128$\\
		&  &UpConv12&$2\times2\times128/2$&$128\times128\times64$\\  
		&  &Concat3&-&$128\times128\times128$\\  
		&  &\textit{ResConv2}&$1\times1\times64/1$& \textit{$128\times128\times64$}\\
		&  &Conv13&$3\times3\times64/1$&$128\times128\times64$\\ 
		&  &Conv14&$3\times3\times64/1$&$128\times128\times64$\\
		&  &Conv15&$3\times3\times64/4$&$128\times128\times4$\\ 
		\hline
		Output & &-&-&$128\times128\times4$\\
		\hline
	\end{tabularx}
\end{table*}

\subsubsection{Image reconstruction network}\label{subsubsec:concat}
The last stage of the proposed TFNet is recovering the desired high resolution MS image from the fused features. The encoded feature maps only take up $\frac{1}{4}\times \frac{1}{4}$ of the input proportion in width and height. The spatial resolution of the feature maps should be up-sampled step by step to meet the resolution of the pan-sharpened image.  One can use traditional method such as linear interpolation to up-sample the feature maps, however learnable way is better~\cite{fcn}. We use backward stride convolutional layers (also known as transposed convolutional layers) to reconstruct the high resolution MS image, 
and up-sample the feature map in every two layers as a symmetrical structure to encoder (i.e., feature extraction and fusion networks). 

It is hard to recover detail textures from high level features, because high layer feature maps encode semantic and abstract information of an image~\cite{gatys2016image}. To recover fine and realistic details, it is better to utilize all levels of representations. However, this will put heavy burden on computation. Inspired by the `crop and copy' operation employed in U-Net~\cite{u-net} that copy features in `contracting path' to that of the corresponding `expansive path', we add \emph{skip connection} between the encoder and decoder of our network\footnote{We prefer to use encoder and decoder to denote feature extraction and image reconstruction, and skip connection to denote bridges between low and high levels of a CNN architecture, because these terms are more frequently used. We do not use crop operation in our network.}. Specifically, after every up-sampling step, the feature maps in the encoder are copied to the decoder and concatenated with the corresponding feature maps to inject more details lost in down-sampling process, as shown in~\cref{fig:architecture}. In experiments, we will discuss how much performance improvement it will be by adding lower features into the higher level. The last layer outputs the desired high resolution 4-band MS image.~\Cref{fig:architecture}(a) shows the detail structure of the TFNet. 

\subsection{Improved with residual blocks}
\begin{figure}[!t]
	\centering  
	\includegraphics[width=0.5\textwidth]{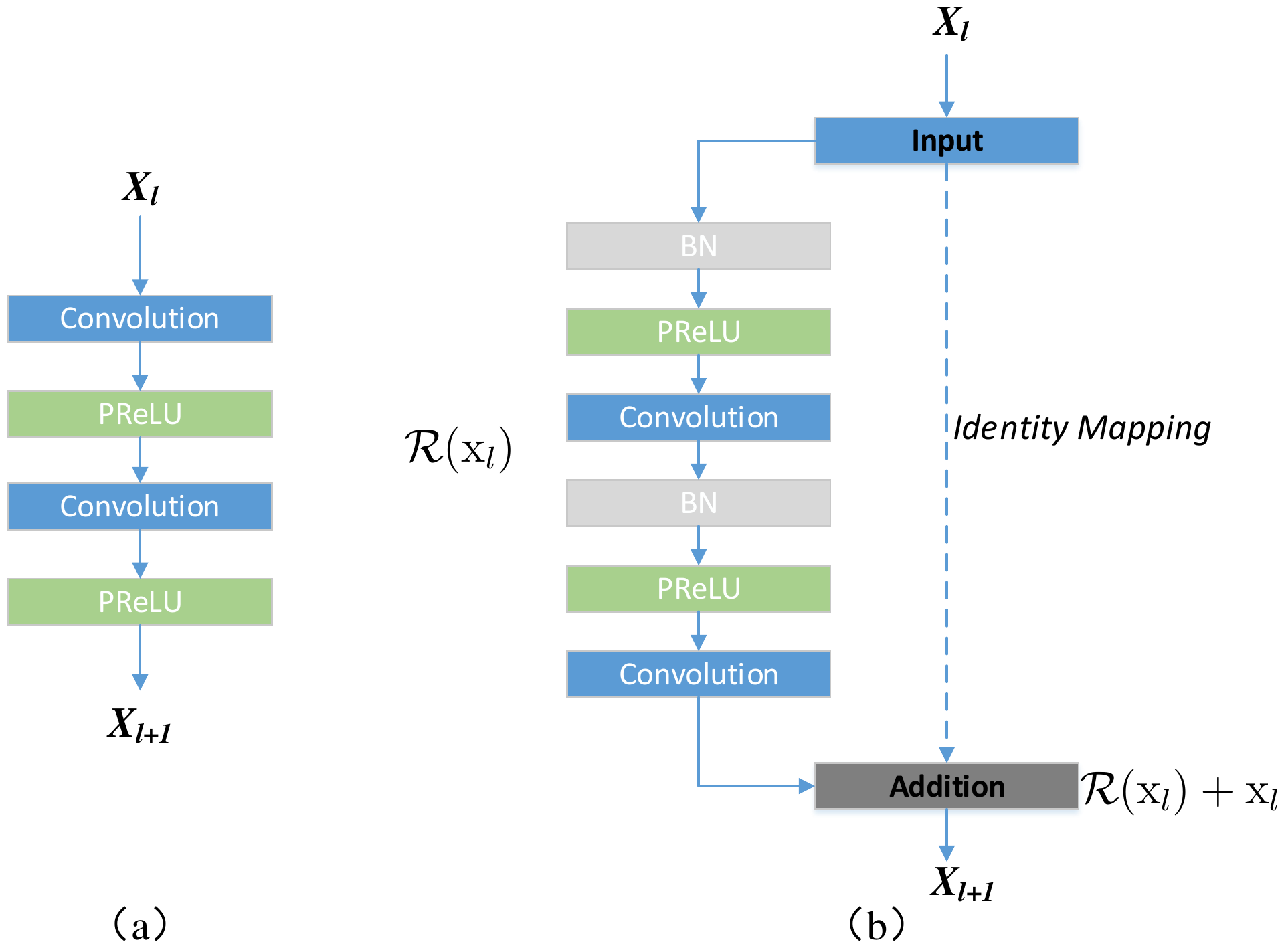} 
	\caption{Structures of different units for building CNN networks. (a) plain unit and (b) residual unit.}
	\label{fig:resblock}  
\end{figure}

\begin{figure*}[!t]
	\centering  
	\includegraphics[width=1\textwidth]{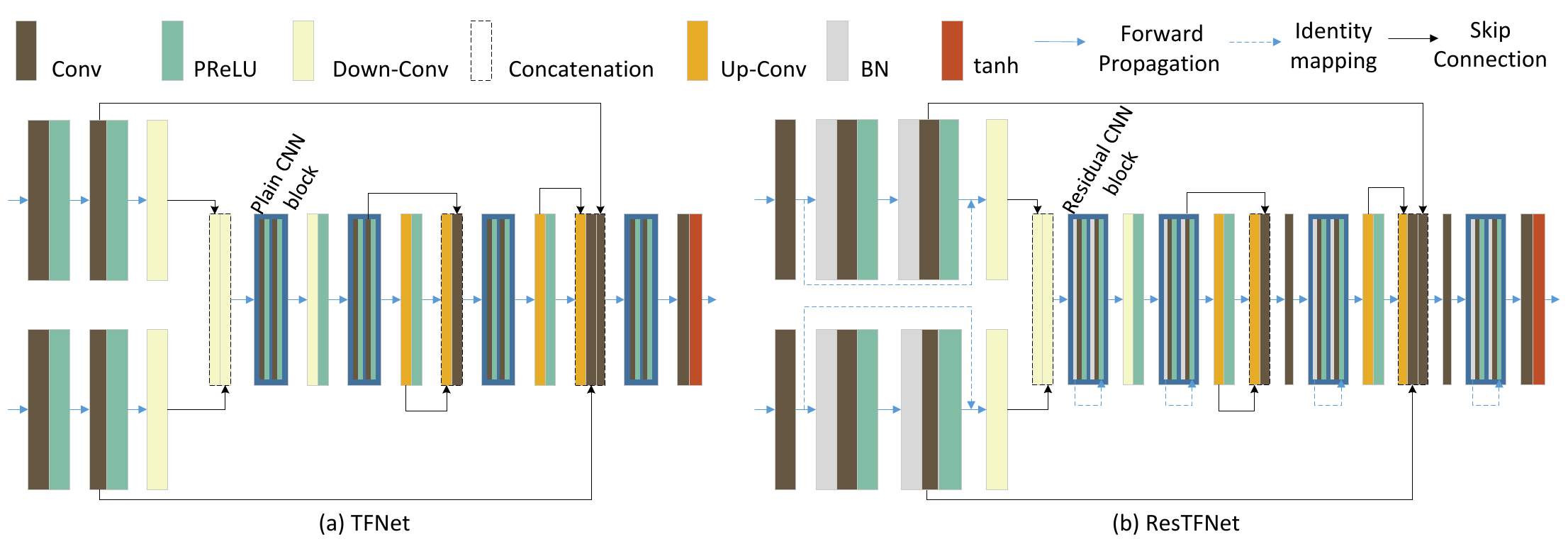} 
	\caption{The detailed architecture of the proposed TFNet and ResTFNet.}
	\label{fig:architecture}  
\end{figure*}

Recent proposed residual learning framework~\cite{he2016deep,he2016identity} has facilitated a lot of vision tasks~\cite{yang2017pannet,lim2017enhanced}. In this work, we also attempt to leverage the power of residual CNN to further improve the performance of our TFNet by replacing the plain CNN units with residual units. Each residual unit can be illustrated as a general form:
\begin{equation}
\begin{split}
\mathbf{y}_{l}\ \ \ & = h(\mathbf{x}_{l})+\mathcal{R}(\mathbf{x}_{l}, \mathcal{W}_{l}), \\
\mathbf{x}_{l+1} & = f(\mathbf{y}_{l}),
\end{split}
\label{Equ:Residual Uint}
\end{equation}
where $\mathbf{x}_{l}$ and $\mathbf{x}_{l+1}$ are the input and output of the $l$-th residual unit, $\mathcal{R}(\cdot)$ is the residual function, $f(\mathbf{y}_l)$ is an activation function and $h(\mathbf{x}_{l})$ is an identity mapping function, a typical one is  $h(\mathbf{x}_{l}) = \mathbf{x}_{l}$. \Cref{fig:resblock} shows the difference between a plain and residual unit. When building residual TFNet (ResTFNet), every two successive convolutional layers are substituted by a residual unit. Because the input and output of a residual unit should have the same size, an additional convolutional layer is added to the behind of each concatenation layer. Thus, the ResTFNet has two more convolutional layers than TFNet. 
The detailed architecture of ResTFNet is shown in~\Cref{fig:architecture}(b) and \Cref{Table:structure} illustrates parameters of the proposed TFNet and its residual variation ResTFNet.

\subsection{Loss function}
The parameters $\Theta$ of TFNet are optimized by minimizing the loss between the pan-sharpened images and the corresponding ground truth images. Besides network architecture, loss function is another important factor affecting  quality of the reconstructed images. Most of the previous restoration tasks employed $\ell_2$ norm as loss function~\cite{dong2016image,kim2016accurate,PNN,yang2017pannet}, however this will produce images suffer from blurring effect. Several studies have suggested that $\ell_1$ loss is a better choice when performing image restoration tasks~\cite{lim2017enhanced,zhao2017loss}, thus in this work, we also prefer $\ell_1$ as loss function to train our network. Given a set of training samples $\{(X^{(i)}_P,X^{(i)}_M,Y^{(i)})\}$, where $X^{(i)}_P,X^{(i)}_M$ are PAN and low resolution MS images, and $Y^{(i)}$ is the corresponding high resolution MS image, $\ell_1$ loss is defines as:
\begin{equation} 
\begin{aligned}
\ell_1(\Theta)=\frac{1}{N}\sum_{i=1}^{N}\left|\Phi\left(X^{(i)}_P,X^{(i)}_M;\Theta \right)-Y^{(i)}\right|_1\\
\end{aligned}
\end{equation}
where $N$ is the number of training samples in a mini-batch.

\section{Experiments and analysis}
\label{sec:experiments}
\subsection{Data sets}
We test the proposed TFNets and compare them with a number of state of the art methods on two image sets collected from Quickbird and GaoFen-1, respectively. Quickbird is a commercial satellite launched on October 18, 2001 by DigitalGlobe\footnote{http://www.digitalglobe.com/}. Quickbird carries two sensors, one sensor acquires PAN images at 0.6 m spatial resolution, and the other sensor acquires 4-band (blue, green, red and near-infrared) MS images at 2.4 m resolution. GaoFen-1, launched on April 26, 2013, is one of a series of high-resolution optical Earth observation satellites of China. It is configured with two 2 m PAN/8 m MS camera and a four 16 m MS medium-resolution and wide-field camera set. In our experiments, only 2 m PAN and 8 m MS data are employed. The spectral wavelength features of GaoFen-1 is similar to that of Quickbird, except that Quickbird possess a higher spatial resolution, as shown in ~\Cref{tb:Spectralandspatialcharacteristic}.
\begin{table*}[!htb]
	\caption{Spectral and spatial characteristics of PAN and MS images for Quickbird and GaoFen-1.}\label{tb:Spectralandspatialcharacteristic}
	\begin{center}
		\renewcommand{\arraystretch}{1.4}
		\setlength\tabcolsep{9pt}
		\begin{tabular}{c|ccccc|cc}
			\hline
			\multirow{2}{*}{Satellite} &\multicolumn{5}{c|}{Spectral Wavelength ($nm$)} &\multicolumn{2}{c}{Spatial Resolution ($m$)}\\
			\cline{2-8}
			& PAN & Blue & Green & Red & Nir & PAN & MS  \\
			\hline
			 Quickbird & 450-900 & 450-520 & 520-600 & 630-690 & 760-890 & 0.6& 2.4 \\
			 GaoFen-1  & 450-900 & 450-520 & 520-590 & 630-690 & 770-890 & 2  & 8 \\
			\hline
		\end{tabular}
	\end{center}
\end{table*}

 For each $w\times h$ MS image, there is a corresponding $4w\times 4h$ PAN image over the same site. The Quickbird dataset contains 9 pairs of MS and PAN images sized from $558\times2080\times4$ to $3162\times2142\times4$ and $2232\times8320\times1$ to $12648\times28568\times1$. And GaoFen-1 dataset is comprised of 4 pairs of MS and PAN images sized from $558\times2080\times4$ to $3162\times2142\times4$ and $2232\times8320\times1$ to $12648\times28568\times1$. Our goal is to generate the MS image with the same spatial resolution to the PAN image. To assess the proposed models, the results should be compared with the referenced images, which do not exist, as a conventional method we follow Wald's protocol~\cite{wald1997fusion} to assess our networks and compare them with other methods. That is down-sampling both PAN and MS images by 4 of width and height separately, then the down-sampled images are used as the inputs of the networks and the original MS images are used as references. We also use bi-cubic interpolation algorithm to up-sample the input MS images to match the PANs' resolution. 

\subsection{Evaluation indexes}
We use six widely used indicators to quantitatively evaluate the performance of the proposed and comparing methods. 
\begin{itemize}
	
\item \textbf{SAM}
The \emph{spectral angle mapper} (SAM)~\cite{Yuhas1992} measures spectral distortions of pan-sharpened image comparing with the reference images. It is defined as angles between the spectral vectors of pan-sharpened and reference image in the same pixel, which can be calculated as:
\begin{equation}\label{equ:sam}
\mathrm{SAM}(\mathrm{x}_1,\mathrm{x}_2)=\arccos\left(\frac{\mathrm{x}_1\cdot \mathrm{x}_2}{\parallel \mathrm{x}_1\parallel \cdot \parallel \mathrm{x}_2\parallel}\right)
\end{equation}
where $\mathrm{x}_{1}$ and $\mathrm{x}_{2}$ are two spectral vectors. SAM is averaged over all the images to generate a global measurement of spectral distortion. For the ideal pan-sharpened images, SAM should be 0.

\item \textbf{CC}
The \emph{Correlation Coefficient} (CC) is another widely used indicator measuring the spectral quality of the pan-sharpened images. It calculates the CC between a pan-sharpened image $X$ and the corresponding reference image $Y$ as
\begin{equation}\label{eq:cc}
\mathrm{CC}=\frac{\sum\limits_{i=1}^{w}\sum\limits^{h}_{j=1}\left(X_{i,j}-\mu_{X}\right)\left(Y_{i,j}-\mu_{Y}\right)}{\sqrt{\sum\limits_{i=1}^{w}\sum\limits^{h}_{j=1}\left(X_{i,j}-\mu_{X}\right)^2\sum\limits^{w}_{i=1}\sum\limits^{h}_{j=1}\left(Y_{i,j}-\mu_{Y}\right)^2}}
\end{equation}
where $w$ and $h$ are the width and height of the images, $\mu_{*}$ indicates mean value of an image. CC ranges from -1 to +1, and the ideal value is +1.

\item \textbf{sCC}
To evaluate the similarity between the spatial details of pan-sharpened images and reference images, a high-pass filter is applied to obtain the high frequencies of them, then the correlation coefficient (CC) between the high frequencies is calculated. This quantity index is called \emph{spatial CC} (sCC)~\cite{zhou1998wavelet}. We use the high Laplacian pass filter given by,
\begin{equation}\label{eq:hpass}
F = \left[
\begin{array}{ccc}
-1 & -1 & -1 \\
-1 & 8 & -1 \\
-1 & -1 & -1
\end{array}
\right]
\end{equation}
to get the high frequency. A higher sCC indicates that most of the spatial information of the PAN image is injected during the fusion process. sCC is computed between each band of the pan-sharpened and reference image. The final sCC is averaged over all the bands of the MS images.

\item \textbf{UIQI}
The \emph{universal image quality index} (UIQI)~\cite{wang2002universal} is an indicator estimating the global spectral quality of the pan-sharpened images. It is defined as
\begin{equation}\label{eq:uiqi}
\mathrm{UIQI}=\frac{\sigma_{XY}}{\sigma_X \sigma_Y}\cdot \frac{2\mu_{X}\mu_{Y}}{ \mu_{X}^2 + \mu_{Y}^2}\cdot \frac{2\sigma_X \sigma_Y}{ \sigma_X^2+ \sigma_Y^2 }
\end{equation}
where $\sigma_{*}$ and $\mu_{*}$ represents standard variance and mean of images, specially $\sigma_{XY}$ is the covariance of image X and Y.

\item {$\mathbf{Q_4}$}
The \emph{Quality-index} Q${_4}$~\cite{alparone2004global} is the 4-band extension of Q index~\cite{wald1997fusion}. Q${_4}$ is defined as:
\begin{equation}
\mathrm{Q_4}=\frac{4|\sigma_{\mathrm{z}_1\mathrm{z}_2}|\cdot|\mu_{\mathrm{z}_1}|\cdot|\mu_{\mathrm{z}_2}|}{(\sigma_{\mathrm{z}_1}^{2}+\sigma_{\mathrm{z}_2}^{2})({\mu_{\mathrm{z}_1}^2}+\mu_{\mathrm{z}_2}^2)}
\end{equation}
where $\mathrm{z}_1$ and $\mathrm{z}_2$ are two quaternions, formed with spectral vectors of MS images, i.e. $\mathrm{z} = a+\mathbf{\mathrm{i}}b+\mathbf{\mathrm{j}}c+\mathbf{\mathrm{k}}d$, $\mu_{\mathrm{z}_1}$ and $\mu_{\mathrm{z}_2}$ are the means of $\mathrm{z}_1$ and $\mathrm{z}_2$, $\sigma_{\mathrm{z}_1\mathrm{z}_2}$ denotes the covariance between $\mathrm{z}_1$ and $\mathrm{z}_2$, and $\sigma_{\mathrm{z}_1}^{2}$ and $\sigma_{\mathrm{z}_2}^{2}$ are the variances of $\mathrm{z}_1$ and $\mathrm{z}_2$.

\item \textbf{ERGAS}
The \emph{erreur relative globale adimensionnelle de synth\`ese} (ERGAS), also known as the relative global dimensional synthesis error is a commonly used global quality index~\cite{Wald2000}. It is given by,
\begin{equation}\label{eq:ergas}
\mathrm{ERGAS} = 100\frac{h}{l}\sqrt{\frac{1}{N}\sum^N_{i=1}\left(\frac{\mathrm{RMSE}(B_i)}{M(B_i)}\right)^2}
\end{equation}
where $h$ and $l$ are the spatial resolution of PAN and MS images; RMSE($B_i$) is the root mean square error between the $i$th band of the fused and reference image; $M(B_i)$ is the mean value of the original MS band $B_i$.
\end{itemize}

\begin{figure*}[!ht]
	\centering
	\includegraphics[width=1\textwidth]{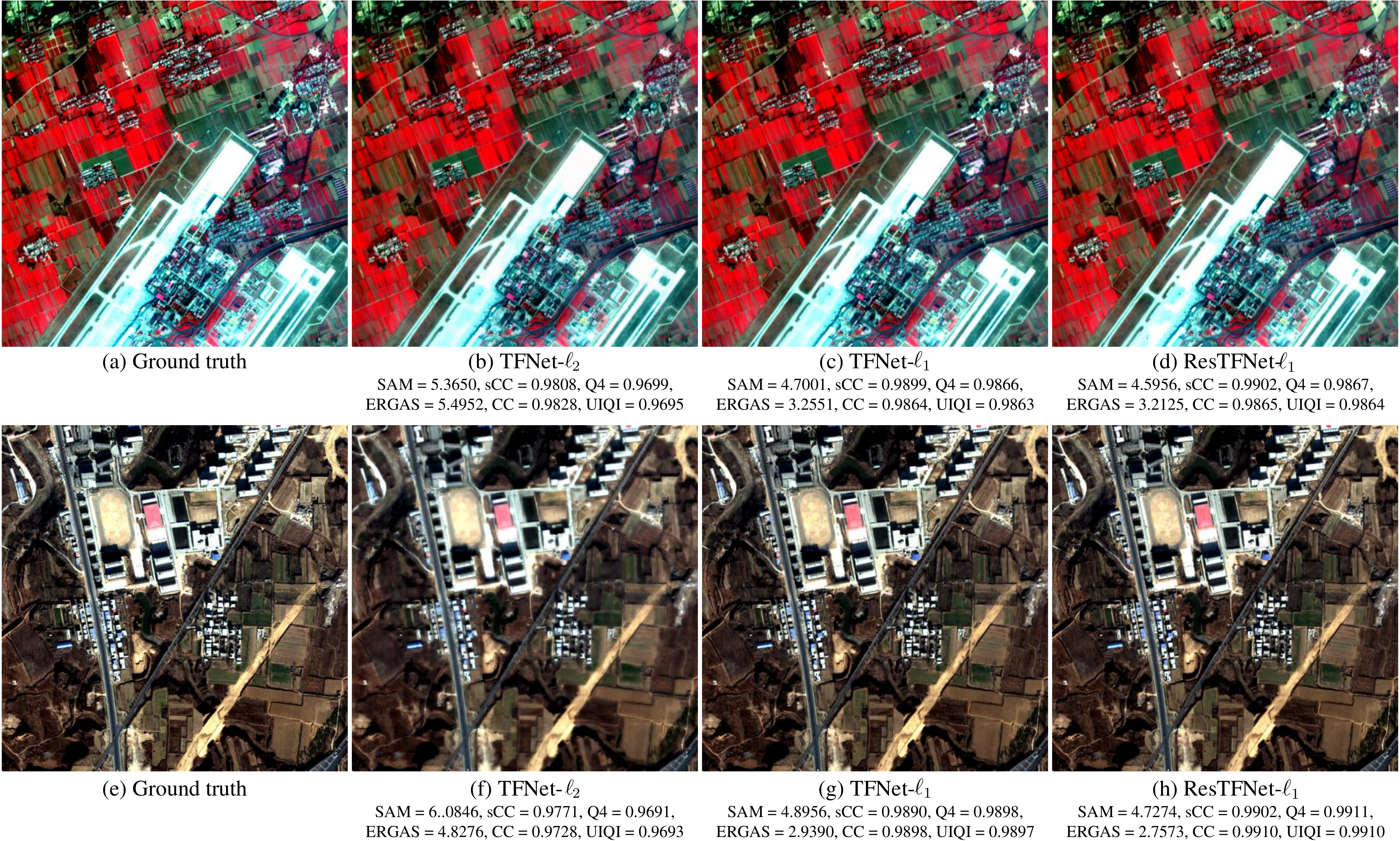} 
	\caption{Comparisons of loss functions and residual learning. The first row are results from GaoFen-1 images and displayed in false color (Nir, red and green bands). The second row are results from Quickbird images and displayed in true color (red, green and blue bands). Better viewed in color.}
	\label{fig:losscomparision}
\end{figure*}

\subsection{Implementation details}
We test our network on Quickbird and GaoFen-1 dataset, separately. In Quickbird dataset, 8 out of 9 images are used to generate training samples and the last one is used as test image. In GaoFen-1 dataset, the number of training and testing images are 3 and 1. We utilize $128\times 128$ images patches to train the TFNets\footnote{Other input sizes such as $256\times256$ or $512\times 512$ are also feasible, however it will need large amount of GPU memory. In this work, we simply use $128\times 128$ images.}. Following Wald's protocol~\cite{wald1997fusion}, these training patches are randomly sampled from the original MS and down-sampled PAN and MS image pairs to form training set. There are 64,000 training samples generated for Quickbird dataset, and 20,000 for GaoFen-1. The network is implemented in PyTorch~\cite{paszke2017automatic} and trained on a NVIDIA Titan X GPU. The loss is minimized using Adam optimizer ~\cite{kingma2014adam} with a learning rate of 0.0001 and a momentum of 0.5. The mini-batch size is set to 32. It will take about 15 hours to train our network. In testing phase, images with size larger than $128\times128$ are firstly divided into image patches with an overlap of 8 before feeding into the TFNets. The final pan-sharpened images are obtained by stitching the outputs together. The pixel values in the overlap regions are averaged.

\subsection{Impacts of detail compensation}

\begin{table*}[!ht]
	\caption{Performance of different concatenation strategies.}\label{tab:concatstrategies}
	\begin{center}
		\renewcommand{\arraystretch}{1.4}	
		\setlength\tabcolsep{3pt}
		\begin{tabular}{l|llllll|llllll}
			\hline
			\multirow{2}{*}{Concatenation} & \multicolumn{6}{c|}{Quickbird}& \multicolumn{6}{c}{GaoFen-1}\\
			\cline{2-13}
			& SAM & CC & sCC & UIQI &Q$_4$ & ERGAS &SAM & CC & sCC & UIQI &Q$_4$ & ERGAS  \\
			\hline
			 None & 4.8831 & 0.9821 & 0.9821 & 0.9818 & 0.9861 & 4.3219 & 4.5074 & 0.9875 & 0.9903 & 0.9873 & 0.9734 & 3.3154 \\
			 PAN  & 4.8601 & 0.9838 & 0.9916 & 0.9838 & 0.9879 & 4.1809 & 4.4347 & 0.9889 & 0.9932 & 0.9889 & 0.9890 & 3.2973 \\
			 PAN+MS & \textbf{4.6811} & \textbf{0.9883} & \textbf{0.9918} & \textbf{0.9882} & \textbf{0.9884} & \textbf{3.9117}  & \textbf{4.4175} & \textbf{0.9891} & \textbf{0.9933} & \textbf{0.9890} & \textbf{0.9891} & \textbf{3.2877} \\		
			\hline
		\end{tabular}
	\end{center}
\end{table*}

In order to reconstruct fine details better, as mentioned in~\cref{subsubsec:concat}, we employ a skip connection strategy by concatenating the low-level feature maps with the corresponding high level feature maps after every up-sampling step. It is widely accepted that PAN images provide spatial details, thus it is natural to compensate the low level details of PAN to the desired pan-sharpened images. An alternative solution is compensating features from both PAN and MS images to the corresponding high level feature maps of the reconstruction network. We evaluate both of these two configurations. Surprisingly, the latter one achieves better performances than the first one in terms of both spatial and spectral qualities, as shown in~\cref{tab:concatstrategies}. The reason behind this may be that both PAN and MS images contain spatial and spectral information. It is impossible to separate spatial and spectral information from each other and extract only spatial details from PAN image or spectral features from MS images. To get better results, one should not focus only on how to extract spatial or spectral information, separately. Both PAN and MS images should be taken into consideration. 

\begin{figure}[!t]
	\centering  
	\includegraphics[width=.5\textwidth]{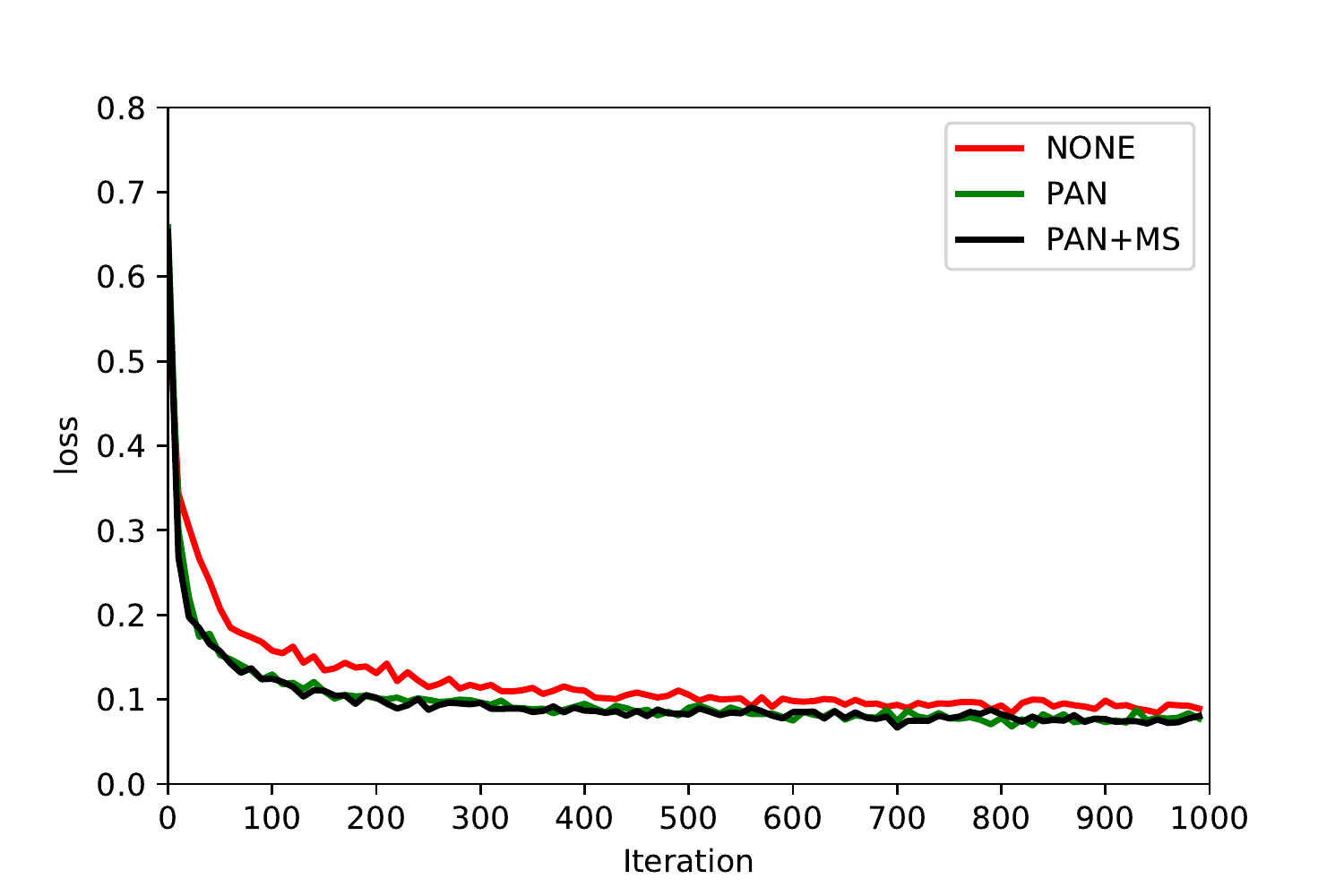} 
	\caption{Loss curves of the proposed TFNet with and without skip connection.}
	\label{fig:loss}  
\end{figure}

Another advantage of adding skip connection is that it could help to alleviate the training problem. The intuition behind this is that copying low level features to the corresponding high levels actually creates a path for information propagation allowing signals propagate between low and high levels in a much easier way, which will facilitate backward propagation during training. As shown in \Cref{fig:loss}, adding skip connection (i.e. compensating details from features of PAN and MS images) not only help to speed up the convergence of the TFNets but also lead to lower training errors. 

\subsection{Evaluation of loss functions and residual learning}

In this experiment, we test both $\ell_1$ and $\ell_2$ losses with the same network architecture. The quantitative measures on the results demonstrate that the TFNet trained with $\ell_1$ (TFNet-$\ell_1$) achieves improved performance than the network trained with $\ell_2$ (TFNet-$\ell_2$). As shown in ~\Cref{fig:losscomparision}(b), (c), (f) and (g), it can be observed that, the pan-sharpened images produced by TFNet-$\ell_1$ shows finer details and sharper edges than that of TFNet-$\ell_2$. The quantitative indicators suggest that $\ell_1$ loss could improve both spatial and spectral qualities of the pan-sharpened images. Especially, the spectral quality is significantly improved (see SAM values of (b) (c), (f) and (g) in \Cref{fig:losscomparision}). Also, the global metrics ERGAS and UIQI of $\ell_1$ are much better than that of $\ell_2$. Replacing plain CNN units with residual units could further boost the performance. As shown in \Cref{fig:losscomparision}(d) and (h). ResTFNet trained with $\ell_1$ (ResTFNet-$\ell_1$) exhibits better results in terms of spatial, spectral and global quantitative metrics. 

\subsection{Comparison with other methods}
\begin{table*}[!ht]
	\caption{Quantitative assessment on the GaoFen-1 image. \mbox{\color{red}{\textbf{Red}}}, \mbox{\color{green}{\textbf{green}}} and \mbox{\color{blue}{\textbf{blue}}} indicate the best, the second best and the third best performance.}\label{tab:gaofen1}	
	\begin{center}
		\renewcommand{\arraystretch}{1.4}
		\setlength\tabcolsep{15pt}
		\begin{tabular}{lllllll}
			\toprule
			\noalign{\smallskip}
			& SAM  & sCC & Q4 & ERGAS & UIQI & CC  \\
			\noalign{\smallskip}
			\midrule
			\noalign{\smallskip}
			AIHS \cite{AIHS}& 12.3300 & 0.8745 & 0.7931 & 13.8698 & 0.7660 & 0.7816 \\ 
			MTF\_GLP\_HPM \cite{aiazzi2003mtf}& 10.0495 & 0.8877 & 0.8083 & 14.1305 & 0.8089 & 0.8183\\ 
			ATWT\_M3~\cite{ranchin2000fusion} & 8.4168 & 0.9461 & 0.9204 & 8.4617 & 0.9189 & 0.9231\\ 
			AWLP \cite{wavelet} & 8.2449 & 0.9241 & 0.8582 & 11.6849 & 0.8568 & 0.8572\\ 
			BDSD \cite{garzelli2008optimal} & 9.0585 & 0.9307 & 0.8895 & 10.229 & 0.8880 & 0.8883 \\ 		
			PNN \cite{PNN}& \color{blue}{\textbf{6.5801}} & \color{blue}{\textbf{0.9838}} & \color{blue}{\textbf{0.9743}} & \color{blue}{\textbf{4.8997}} & \color{blue}{\textbf{0.9741}}& \color{blue}{\textbf{0.9746}} \\
			TFNet$-\ell_1$ & \color{green}{\textbf{4.4175}} & \color{green}{\textbf{0.9933}} & \color{green}{\textbf{0.9891}} & \color{green}{\textbf{3.2877}} & \color{green}{\textbf{0.9890}}& \color{green}{\textbf{0.9891}}\\
			ResTFNet$-\ell_1$ & \color{red}{\textbf{4.3316}} & \color{red}{0.9935} & \color{red}{\textbf{0.9896}} & \color{red}{\textbf{3.2011}} & \color{red}{\textbf{0.9896}}& \color{red}{\textbf{0.9895}}\\
			\bottomrule
		\end{tabular}
	\end{center}
\end{table*}

\begin{table*}[!ht]
	\caption{Quantitative assessment on the Quickbird image. \mbox{\color{red}{\textbf{Red}}}, \mbox{\color{green}{\textbf{green}}} and \mbox{\color{blue}{\textbf{blue}}} indicate the best, the second best and the third best performance.}\label{tab:qb}	
	\begin{center}
		\renewcommand{\arraystretch}{1.4}
		\setlength\tabcolsep{15pt}
		\begin{tabular}{lllllll}
			\toprule
			\noalign{\smallskip}
			& SAM  & sCC & Q4 & ERGAS & UIQI & CC  \\
			\noalign{\smallskip}
			\midrule
			\noalign{\smallskip}
			AIHS \cite{AIHS}& 7.8209 & 0.9571 & 0.9401 & 8.3139 & 0.9382& 0.9539 \\ 
			MTF\_GLP\_HPM \cite{aiazzi2003mtf}& 8.5199 & 0.9557 & 0.9580 & 7.5465 & 0.9557 & 0.9568\\ 
			ATWT\_M3~\cite{ranchin2000fusion} & 7.1677 & 0.9242 & 0.8692 & 12.9358 & 0.8626 & 0.8776\\ 
			AWLP \cite{wavelet} & 7.2995 & 0.9478 & 0.9401 & 9.5812 & 0.9236 & 0.9288\\ 
			BDSD \cite{garzelli2008optimal} & 6.5661 & 0.9634 & 0.9591 & 7.3583 & 0.9576 & 0.9582\\ 		
			PNN \cite{PNN}& \color{blue}{\textbf{6.3484}} & \color{blue}{\textbf{0.9829}} & \color{blue}{\textbf{0.9679}} & \color{blue}{\textbf{6.2825}} & \color{blue}{\textbf{0.9668}}& \color{blue}{\textbf{0.9704}}\\
			TFNet$-\ell_1$ & \color{green}{\textbf{4.6811}} &\color{green}{\textbf{0.9918}} & \color{green}{\textbf{0.9884}} & \color{green}{\textbf{3.9117}} & \color{green}{\textbf{0.9882}}& \color{green}{\textbf{0.9883}}\\
			ResTFNet$-\ell_1$ & \color{red}{\textbf{4.5149}} & \color{red}{\textbf{0.9925}} &\color{red}{\textbf{0.9896}}  &\color{red}{\textbf{3.6931}}  & \color{red}{\textbf{0.9895}}& \color{red}{\textbf{0.9896}}\\
			\bottomrule
		\end{tabular}
	\end{center}
\end{table*}
In this subsection, we compare the proposed method with several widely used techniques, including IHS based method~\cite{carper1990use}, additive wavelet luminance proportional (AWLP) method~\cite{wavelet}, \`a trous wavelet transform with the injection model 3 (ATWT\_M3)~\cite{ranchin2000fusion}, generalized Laplacian pyramid with MTF-matched filter and multiplicative injection model (\mbox{MTF\_GLP\_HPM})~\cite{aiazzi2003mtf}, band-dependent spatial-detail with local parameter estimation (BDSD) method~\cite{garzelli2008optimal}, and CNN based pan-sharpening (PNN)~\cite{PNN}. 

\begin{figure*}[!htb]
	\centering  
	\includegraphics[height=.95\textheight]{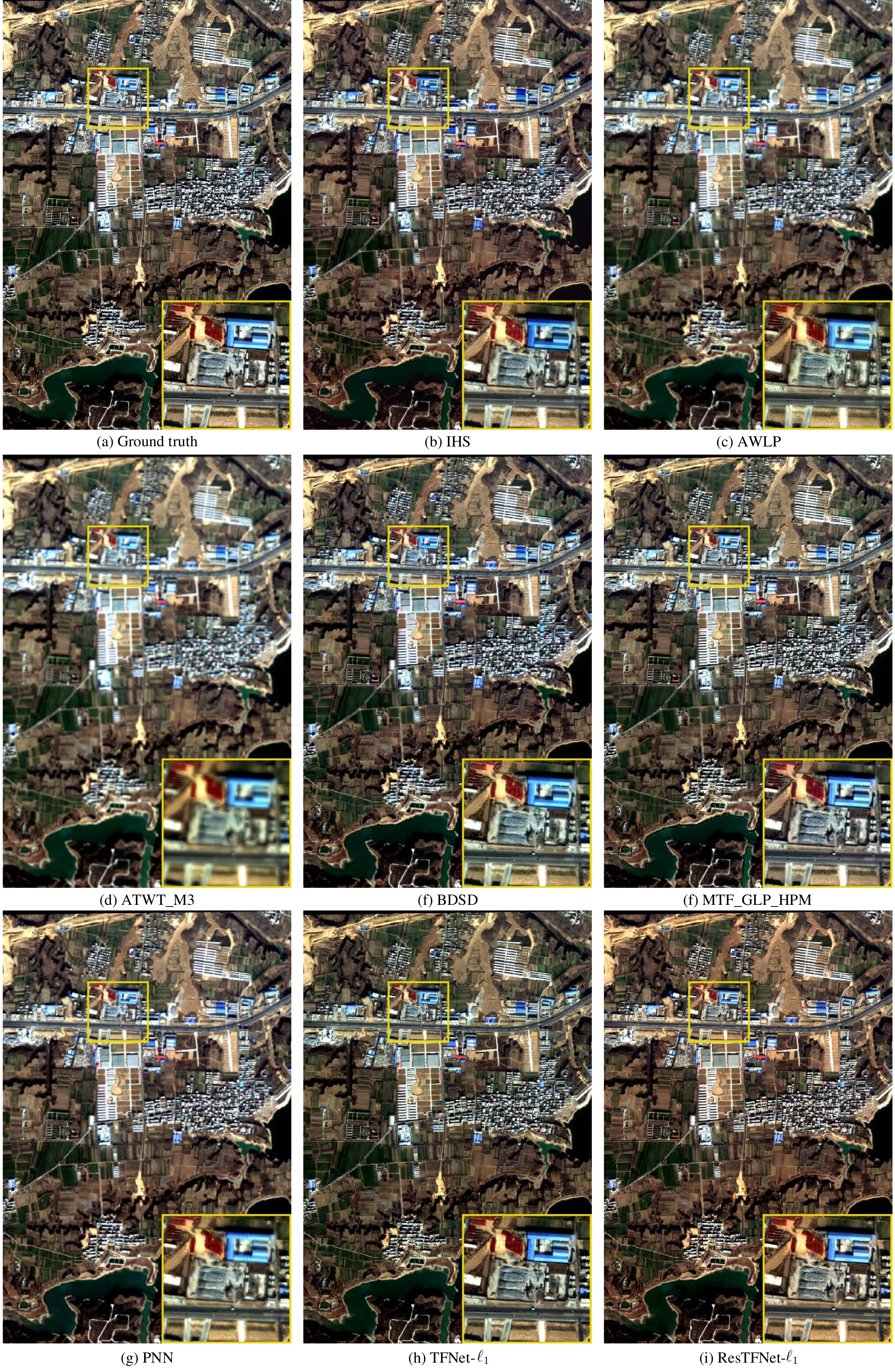} 
	\caption{Pansharpening results on the Quickbird images.}
	\label{fig:compareonqb}  
\end{figure*}

\begin{figure*}[!htb]
	\centering  
	\includegraphics[width=1\textwidth]{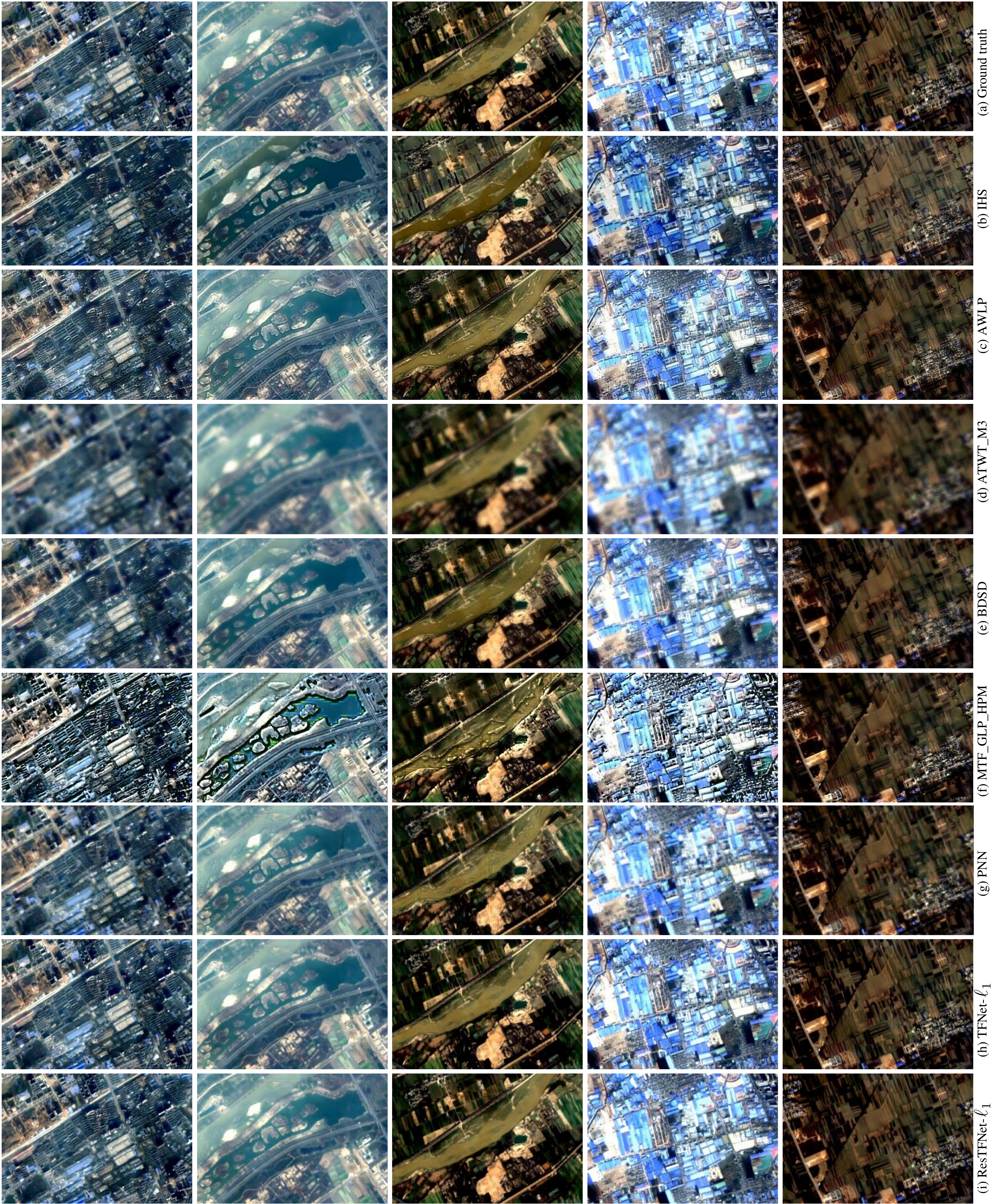} 
	\caption{Sub-regions of the results on the GaoFen-1 image. The first row is the ground truth images. Row 2 to 8 present the results of the comparing methods and the proposed TFNet and ResTFNet.}
	\label{fig:comparegf}  
\end{figure*}

\Cref{tab:qb,tab:gaofen1} report the quantitative metrics on the two test images. \Cref{fig:compareonqb,fig:comparegf} illustrate example regions cropped from the pan-sharpened images. From \Cref{tab:qb,tab:gaofen1} we can see that the proposed TFNets, including TFNet-$\ell_1$ and ResTFNet-$\ell_1$ obtain the best performance in term of all the indicators. Building the network with residual units could further improve the qualities of the pan-sharpened images, especially the spectral distortions can be reduced, significantly. PNN~\cite{PNN} method also achieves good results on the both test images, in which all the indicators surpass traditional methods. The results show great potential of deep learning methods in solving pan-sharpening problems. Besides the proposed TFNets and PNN, BDSD~\cite{garzelli2008optimal} and MTF\_GLP\_HPM~\cite{aiazzi2003mtf} obtain very promising results. This is consistent with Vivone et al.'s study~\cite{vivone2015critical}.

From \Cref{fig:compareonqb} we can see that all the methods produce visually satisfactory pan-sharpened images, except ATWT\_M3 and PNN methods. ATWT\_M3 method suffer from very severe blurring and artifacts. PNN method also presents blurring effects, as it can be easily seen from \Cref{fig:compareonqb}(g), some missing spatial details are noticeable. Obvious spectral distortions can be identified from IHS method (see \Cref{fig:compareonqb}(b)). The color of the rooftop is darker than the reference image. \Cref{fig:comparegf} shows sub sets of the results on the GaoFen-1 test image. It can be clearly observed that the MTF\_GLP\_HPM method has severe spectral distortions and aliasing, especially in water regions, see the second and third columns of \Cref{fig:comparegf}(f). Although IHS method produces images with good visual quality, spectral distortions are also noticeable. Similar to the results on Quickbird images, blurring effects occur again in the GaoFen-1 results of ATWT\_M3 and PNN methods. Both AWLP and BDSD generate pan-sharpened images with reduced spectral distortions, however there exist blurring effects. The proposed methods does better in spectral preservation and provides images with richer spatial details. 
%

\section{Conclusion}
\label{sec:conclusion}
In this paper, we have proposed a two-stream fusion network for solving remote sensing image fusion, i.e. pan-sharpening problem. The proposed TFNets was motivated by the recent progresses achieved by deep learning techniques, especially CNN architectures. A flexible and powerful tool it is, CNN can extract hierarchical features of an input image. One can easily design CNNs to extract features of PAN and MS images. Then, it is naturally to fuse features of them to accomplish pan-sharpening. Different from the previous CNN based methods that perform pan-sharpening in pixel level, we explore to fuse PAN and MS images in feature domain. The experiments on Quickbird and GaoFen-1 images demonstrate that the proposed method can produce pan-sharpened images with promising spatial and spectral qualities. The results can be further improved through $\ell_1$ and residual learning. As a future work, we will develop loss functions more suitable for pan-sharpening task and study unsupervised methods. 
\section{Acknowledgements}
This work is supported by the National Natural Science Foundation of China under the Grant 61601011, and in part by National Major Program on High Resolution Earth Observation System under the Grant 591 03-Y30B06-9001-13/15-01.
\section*{References}
\bibliographystyle{elsarticle-num}
\bibliography{mybib}

\end{document}